\documentclass[11pt]{article}
\usepackage{amssymb,amsbsy,amsfonts,amsmath,amsthm,xspace}
\usepackage{mathrsfs}
\usepackage{graphicx}
\usepackage{caption}
\usepackage{setspace}
\usepackage{subcaption}
\usepackage{mymacros}
\usepackage{multirow}
\usepackage{epstopdf}
\usepackage{epsfig}
\usepackage{url}
\usepackage[toc,page]{appendix}
\usepackage{float}
\usepackage{natbib}
\usepackage{color}
\usepackage{verbatim}

\begin{document}

\def\text#1{\mbox{\rm #1}}
\def\overset#1#2{\stackrel{#1}{#2} }

\def\mb{\mathbf}
\def\mr{\mathrm}
\def\dsum{\displaystyle\sum}
\def\dint{\displaystyle\int}
\def\dfrac{\displaystyle\frac}

\renewcommand{\baselinestretch}{1}
%
%
\newcommand{\pkg}[1]{\textsf{#1}}

\title{\LARGE \bf A note on the statistical view of matrix completion}

\author{Tianxi Li \\ \normalsize Department of Statistics,\\
 \normalsize University of Michigan, Ann Arbor \\}

\maketitle
\bigskip

\begin{abstract}
A very simple interpretation of matrix completion problem is introduced based on statistical models. Combined with the well-known results from missing data analysis, such interpretation indicates that matrix completion is still a valid and principled estimation procedure even without the missing completely at random (MCAR) assumption, which almost all of the current theoretical studies of matrix completion assume.
\end{abstract}

\section{Introduction}
Matrix completion has attracted a great amount of attention in the past ten years \cite{candes2009exact, candes2010power, candes2010matrix, mazumder2010spectral, keshavan2009matrix, davenport20141, klopp2015matrix}. A low-rank assumption is made to ensure a matrix, though with a larger proportion of entries being unobserved, to be estimable in certain senses. Given a matrix $Y \in \bR^{m\times n}$ with missing entries, we denote the index set of observed positions as $\Omega \subset [m]\times[n]$ and define the projection operator $P_{\Omega}$ as a mapping from $\bR^{m\times n}$ to $\bR^{m\times n}$, such that
$$P_{\Omega}A  = (A_{ij}I((i,j)\in \Omega))_{m\times n},$$
where by convention, we define a missing value multiplied by zero is still zero. With these notations, the most straightforward way of formulating a low-rank matrix completion procedure is 
\begin{equation}\label{eq:SVD-lowrank}
\ourmin_{\Rank(Z)=q}\norm{P_{\Omega}Y - P_{\Omega}Z}_F^2,
\end{equation}
where $\norm{\cdot}_F$ is the Frobenius norm. This is a very natural generalization of the classical SVD problem to the situation where missing entries are present, so we call this problem missing value SVD. Missing value SVD is non-convex, due to the constraint on rank, and currently there is no efficient algorithm that guarantees global optimal solution. One approach to deal with the non-convexity is taking a convex relaxation of the rank. The most popular relaxation currently used is the nuclear norm $\norm{Z}_*$, which is the sum of singular values of $Z$. For instance, \cite{mazumder2010spectral} takes the Langrangian problem  
\begin{equation}\label{eq:nuclear}
\ourmin_{Z}\norm{P_{\Omega}Y - P_{\Omega}Z}_F^2 + \lambda\norm{Z}_*.
\end{equation}

Other similar formulations are also available \cite{candes2010matrix, davenport20141}. \newline

Now we proceed to introduce a few terminologies of missing data mechanism in statistics from \cite{little2014statistical}. Let $M$ be the indicator matrix of missing positions, such that $M_{ij}=1$ if and only if $(i,j) \notin \Omega$. Then we say the missing mechanism is {\em missing completely at random (MCAR)} if
$$\p(M|Y,\phi) = \p(M|\phi)$$
where $\phi$ is certain unknown underlying parameters of the missing process. For example, in the case of uniformly missing, $\phi$ can be the probability that each entry is missing. A more general mechanism is called {\em missing at random (MAR)} which assumes the missing indicator only depends on the observed values as
$$\p(M|Y,\phi) = \p(M|\{Y_{ij}\}_{(i,j)\in \Omega},\phi).$$
If the missing indicator also has dependence on the missing entries, given all the observed ones, then it is called {\em not missing at random (NMAR)}. To the best of our knowledge, all of the theoretical results about matrix completion are assuming that the missing positions are MCAR. Such assumption is not realistic in many applications. For instance, in the Netflix problem where the missing entries are the movie scores that users never watched, MCAR is clearly unrealistic (while MAR may be better but still a bit too strong). In next section, we will treat the matrix completion as a statistical model estimation problem and show that the matrix completion is valid for MAR.

\section{Statistical models for matrix completion}

Assume $Y = Z +E$, where $Z$ is a parameter matrix of rank $q$. $E = (e_{ij})_{m\times n}$ such that $e_{ij}$'s are i.i.d $\ncal(0,\sigma^2)$. $\sigma^2$ can be assumed to be known without loss of generality. It is easy to see that the log likelihood kernel of the model is
$$\ell(Z;Y) \propto \exp(\frac{1}{2\sigma^2}\sum_{ij}(Y_{ij}-Z_{ij})^2).$$

Therefore, using the fact that the rank-$q$ SVD is the rank-$q$ matrix that has the smallest Frobenius error, we know that the rank-$q$ SVD is the MLE of $Z$. Using the same idea, we can see that the missing-value SVD is actually trying to solve the MLE for the log likelihood

$$\ell(Z;Y_{ij}, (i,j)\in \Omega).$$
This is the likelihood after integrating out all of the missing entries in the model. According to \cite{little2014statistical}, we just need the following ignorable assumption the make such MLE is a valid estimate.
\begin{ass}[Ignorable assumption 1]
The missing mechanism is MAR and the model parameter space for $(Z, \phi)$ is a product space of sets for $Z$ and $\phi$.
\end{ass}

The second half of the assumption is trivial, thus the ignorable assumption essentially indicates that MAR is enough for a valid estimation. On the other hand, problem \eqref{eq:nuclear} can be seen as an approximate MLE to the ignorable likelihood. However, it will also be interesting to see what is the statistical interpretation by itself, which may give a better view of the needed missing mechanism. We now define the following Bayesian model:
Let $O(m,n)$ be the space of $m \times n$ orthogonal matrices. 
\begin{itemize}
\item Noninformative improper prior $U \sim Unif(O(m,q))$, $V \sim Unif(O(n,q))$.
\item $D = \diag(\theta)$ and $\theta_k$'s are i.i.d from a Laplace distribution $\lcal(0,b)$, $k=1, \cdots, q$.
\item $Y|U,D,V  = UDV^T + E$, $e_{ij}$ i.i.d $\ncal(0,\sigma^2)$.
\item $M|Y,\phi \sim \p(M|Y,\phi)$.
\end{itemize}
Now suppose in the likelihood, we integrate the missing positions out and use the resulting marginal likelihood of $Y_{ij}, (i,j)\in \Omega|U,D,V$ in the Bayes estimation. It is not hard to see that the problem for solving the posterior mode is the following one:

$$\ourmin \sum_{(i,j)\in \Omega}(Y_{ij}-Z_{ij})^2 + \lambda\norm{Z}_*$$

which is exactly problem \eqref{eq:nuclear}. To make the estimation valid, we just need a Bayesian ignorable assumption as discussed in \cite{little2014statistical}:

\begin{ass}[Ignorable assumption 2]
The missing mechanism is MAR and the priors for $Z$ and $\phi$ are independent.
\end{ass}

The second part is trivial thus the condition we need is still MAR.

\section{Simulation example of matrix completion under MCAR and MAR}

In this example, we generate data with MCAR and MAR then compare the performance of  \eqref{eq:nuclear} in the two situations. In simulation, we constrain that MAR data has the same amount of missing entries and similar missing position distributions as with the MCAR data, up to permutation of rows. So the performance should be purely about the power under the two mechanisms. The performance is measured by the average relative errors over 100 replications. The relative error is defined as

$$\sum_{(i,j)\in \Omega}(Z_{ij}-\hat{Z}_{ij})^2/\norm{Z}_F^2 \times 100\%.$$
The performance on $300 \times 300$ matrices is given in Table 1. As can be seen that the performance under MCAR and MAR are nearly the same. This can also be checked by two-sample tests which gives large p-values. Essentially, it shows that matrix completion still works well under MAR.

\begin{table}[ht]
\centering

\begin{tabular}{r|r|rrrr}
  \hline
Rank & mechanism & 10\% & 30\% & 50\% & 80\% \\ 
  \hline
\multirow{2}{*}{ 5}& MAR & 0.0538 & 0.1644 & 0.2838 & 0.5512 \\ 
&MCAR & 0.0539 & 0.1641 & 0.2822 & 0.5535 \\ 
  \hline
\multirow{2}{*}{20}& MAR & 0.0626 & 0.1985 & 0.3630 & 1.3950 \\ 
& MCAR & 0.0627 & 0.1975 & 0.3645 & 1.4018 \\ 
   \hline
\end{tabular}
\label{tab1}
\caption{Relative imputation errors for different missing proportions and ranks. The performance differences between the two missing mechanism are very small.}
\end{table}

\section{Summary}

We draw statistical interpretations of two most popular matrix completion techniques. Such statistical interpretations reveal that MAR missing mechanism is enough for matrix completion procedure to be a valid statistical estimation. Though assuming MCAR makes it easier for deriving theoretical properties of matrix completion in general, such condition may be unnecessary for good performances in practice.

\bibliography{tianxibib}{}
\bibliographystyle{abbrv}

\end{document}